PAPER

# High-Fidelity 3D Geometric Reconstruction of Pelvic Organs from MRI: A Hybrid Deep Learning and Iterative Optimization Approach

**Hui Wang[1,2,#], Xiaowei Li[3,#], Chenxin Zhang[1,2], Yifan Feng[3], Jianwei Zuo[1,2], Yumeng Tang[1,2], Xiuli Sun[3], Jianliu Wang[3], Bing Xie[3,*] and Jiajia Luo[1,2,*]**

[1] Institute of Medical Technology, Peking University Health Science Center, Peking University, Beijing, People's Republic of China
[2] Biomedical Engineering Department, Institute of Advanced Clinical Medicine, Peking University, Beijing, People's Republic of China
[3] Department of Obstetrics and Gynecology, Peking University People's Hospital, People's Republic of China

[#]Equal contribution.
*Author to whom any correspondence should be addressed.

**E-mail:** bingxie@bjmu.edu.cn and jiajia.luo@pku.edu.cn

## Abstract

*Objective.* Patient-specific three-dimensional (3D) reconstruction of pelvic organ geometry from pelvic magnetic resonance imaging (MRI) is important for pelvic floor modeling and downstream patient-specific analysis. However, while previous studies have focused primarily on either image segmentation or downstream use of 3D models, the reconstruction of high-fidelity, high-quality geometries remains labor-intensive and poorly standardized.

*Approach.* The study introduced a hybrid deformable shape modeling framework that integrates deep learning prediction with iterative optimization for the reconstruction of the bladder, uterus, and rectum. The framework consists of three core components: (1) a geometry-aware multi-level deep learning architecture that preserves topological consistency of pelvic organs; (2) a two-stage amortized optimization training strategy that balances global shape capture and local surface refinement; and (3) a holistic synergy mechanism—where iterative optimization provides supervision for deep learning during the training phase, and during inference, deep learning rapidly predicts the global organ morphology, followed by iterative optimization to refine local surfaces and mesh quality.

*Main results.* This framework demonstrated marked superiority in geometric fidelity compared with current mainstream deep learning-based organ reconstruction models. For individual anatomical structures, the reconstructed 3D geometries for the bladder, rectum, and uterus achieved significantly lower Chamfer Distance values and higher Dice Similarity Coefficient scores. In addition, while maintaining high computational efficiency, the proposed architecture yielded superior overall volumetric mesh quality. At the patient level, the framework achieved higher mean values for the 10 worst elements for both minSICN and minSIGE compared to traditional geometric post-processing algorithms.

*Significance.* We present an efficient 3D geometric reconstruction framework that successfully bridges the gap between image-domain segmentation and structural-domain analysis in pelvic floor imaging research. By enabling rapid generation of high-fidelity, high-quality geometries, this framework provides key support for patient-specific pelvic floor modeling and analysis in pelvic organ prolapse.

## 1. Introduction

Pelvic organ prolapse (POP) is a prevalent gynecological disorder characterized by anatomical displacement of pelvic organs from weakened pelvic floor support (Jelovsek *et al* 2007). Driven by global population aging, the incidence of POP continues to rise (Wu *et al* 2009). In the United States alone, approximately 200,000 women undergo surgical intervention annually, with associated healthcare expenditures exceeding $1 billion each year (Boyles *et al* 2003, Subak 2001). This condition significantly compromises both physical function and overall quality of life in affected women.

Magnetic resonance imaging (MRI), by virtue of its excellent soft-tissue contrast and the absence of ionizing radiation, has become an important diagnostic modality for evaluating POP (Woodfield *et al* 2010). However, conventional MRI sequences provide only a limited representation of the intricate three-dimensional (3D) spatial relationships within the pelvic floor, hindering the precise quantification of organ deformation, support defects, and complex spatial interactions.

Consequently, high-fidelity 3D geometric reconstruction of pelvic organs from MRI has become increasingly relevant for patient-specific pelvic floor modeling and analysis. These models facilitate objective quantification of anatomical heterogeneity, supporting morphological assessment and etiological analysis (Larson *et al* 2012a, Hong *et al* 2023, Martin *et al* 2025, Larson *et al* 2012b). Furthermore, they serve as an essential prerequisite for patient-specific biomechanical simulations (Luo *et al* 2015, Gordon *et al* 2019). By leveraging 3D geometries to construct finite element representations of the pelvic floor, clinicians can further obtain critical quantitative evidence to predict disease progression, optimize individualized surgical planning, and evaluate postoperative outcomes.

In conventional workflows, constructing these 3D models remains a highly labor-intensive and operator-dependent process. Specialized radiologists are required to perform slice-by-slice manual or semi-automated segmentation, followed by time-consuming geometric refinement and post-processing. Given the anatomical complexity of the pelvic floor and substantial inter-patient variability, generating high-quality 3D geometries demands extensive expertise, taking hours to days per case. Moreover, operator subjectivity inevitably introduces inter-observer variability in reconstruction outcomes. This low-efficiency, high-burden, and poorly standardized workflow substantially limits the broader use of patient-specific pelvic floor modeling and analysis in larger clinical and research cohorts (Hoyte and Damaser 2016).

Deep learning (DL)-driven analytical tools have reached a high level of maturity within the domain of MRI, demonstrating exceptional efficacy in semantic segmentation and automated diagnosis across diverse anatomical regions (Ma *et al* 2024, Sun *et al* 2024). Within the domain of pelvic floor medicine, research is also progressively advancing. These AI-based techniques have been extended to include automated anatomical segmentation and precision landmark localization (Feng *et al* 2020). Furthermore, segmentation robustness under data-scarce conditions has been enhanced through semi-supervised learning paradigms (Zuo *et al* 2026).

Despite these advancements, current research paradigms remain predominantly confined to voxel-wise image processing. While achieving high voxel-level accuracy for pelvic organ segmentation (Kuş and Aydin 2024, Isensee *et al* 2021, Ma *et al* 2024) , the resulting segmentation masks often suffer from rough surfaces, non-physiological distortions, and poor structural smoothness when converted into 3D geometries due to the lack of explicit 3D geometric and topological constraints (He *et al* 2019, Feng *et al* 2020, Zuo *et al* 2026).

Consequently, these geometries often require extensive manual repair involving complex geometric processing within specialized software, which substantially limits workflow efficiency and standardization. The transition from discrete segmentation masks to continuous, high-fidelity 3D geometries—essential for clinical research and biomechanical modeling—still necessitates labor-intensive and often inconsistent post-processing workflows. This "technical gap" is particularly pronounced in biomechanical simulations, which impose stringent requirements on the topological integrity and surface smoothness of the mesh representations (Sbriglio *et al* 2026, Feng *et al* 2025).

To address these issues, DL-based 3D geometric reconstruction has been developed to map 2D images or 2D segmentation masks to continuous and anatomically consistent 3D organ shapes. By learning the nonlinear mapping between image features and spatial geometry, this paradigm can alleviate topological discontinuities and geometric abnormalities with anatomical priors (Chen *et al* 2024, Lin *et al* 2025, McMillian *et al* 2025, Kong *et al* 2021, Beetz *et al* 2023, Meng *et al* 2024, Laumer *et al* 2025, Ma *et al* 2023, Bongratz *et al* 2024, Hu *et al* 2024, Qian *et al* 2025). Nevertheless, direct application to the pelvic floor remains challenging due to large inter-individual anatomical variations and the extreme scarcity of high-quality 3D annotated datasets, leading to limited generalization of existing DL models.

Therefore, in this work, we aimed to develop a hybrid framework specifically tailored for high-quality 3D geometric reconstruction of pelvic organs. Our proposed method integrates deep learning with iterative optimization to overcome the bottleneck of rapidly synthesizing high-fidelity 3D pelvic organ geometries with improved mesh-quality proxy performance, starting from discrete mask representations. The main contributions of this study are as follows:

- We propose a geometry-aware multi-level deep learning architecture that fuses dual-branch graph features via geometric convolution and cross-attention, preserving topological consistency while capturing global and fine-grained geometry from point clouds.
- We develop a two-stage amortized optimization training strategy that first learns global shape via supervised full amortization, then refines geometric quality via semi-amortization, decoupling reconstruction fidelity and quality constraints to enhance training stability.
- We introduce a synergistic mechanism between deep learning and iterative optimization, where iterative optimization supervises training via pseudo-labels, and deep learning provides hot-start initialization for rapid iterative refinement during inference, achieving both high fidelity and efficiency.

## 2. Materials and Methods

### 2.1. Datasets

The study cohort comprised 100 female participants, with a total of 1,994 MR images analyzed. Each subject underwent sagittal (SAG) MR imaging using a T2-weighted fast spin-echo sequence (repetition time, 3072 ms; echo time, 70 ms; slice thickness, 4 mm; slice gap, 1 mm; field of view, 24×24 cm; and matrix, 344×262). The MR data were acquired at Peking University People's Hospital using a 3T scanner (Philips Healthcare, Bothell, WA, USA) and the study was approved by the institutional ethics committee (#2020PHB092-01). Considering the unique anatomical challenges of the female pelvic floor, this study focuses on three clinically relevant organs in female pelvic MRI: the bladder, the uterus, and the rectum.

### 2.2. Hybrid deformable shape modeling framework

We developed a hybrid deformable shape modeling framework that integrates DL-based prediction with an iterative optimization module. As illustrated in Figure 1, the architecture comprises four sequential stages:

1) Initial geometry extraction and canonical normalization: Target organ regions are first delineated from MRI-derived segmentation masks (Feng *et al* 2020). Subsequently, the Marching Cubes algorithm (Lorensen and Cline 1987) is employed to generate initial geometries from the segmentation masks.

2) DL-driven deformation prediction: This stage focuses on learning the complex nonlinear morphological mapping between segmentation-derived geometry and organ morphology using deep neural networks. The network extracts point cloud features and predicts a global displacement field that deforms a canonical spherical template (Wang *et al* 2021) into organ-specific target morphologies, thereby enabling rapid construction of anatomically coherent 3D geometries.

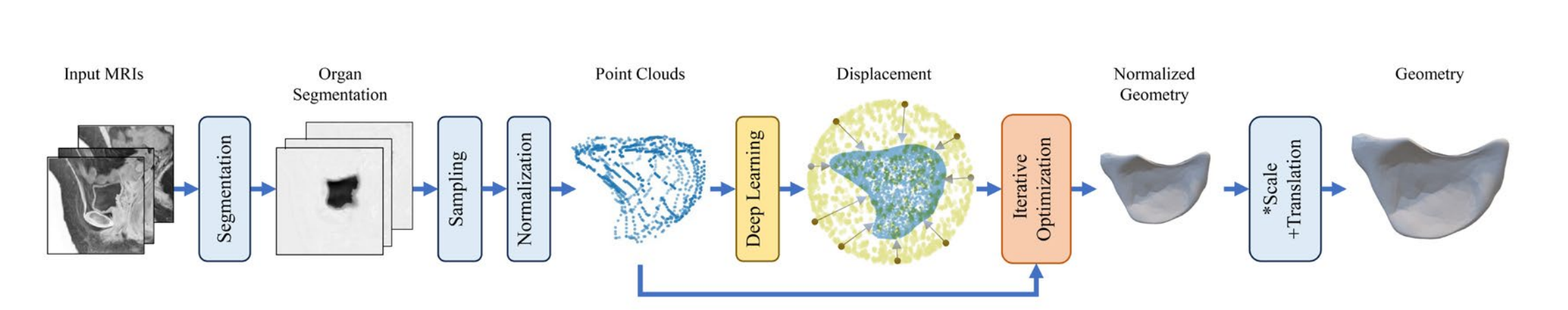

**Figure 1.** Overview of the hybrid framework for pelvic organ 3D geometric reconstruction. From left to right, pelvic MR images are segmented to obtain organ masks, which are then sampled and normalized into point clouds for deep learning-based displacement prediction. The predicted geometry is further refined by iterative optimization, followed by inverse scale transformation to generate the final patient-specific geometry.

3) Geometry-constrained iterative refinement: While DL effectively captures global organ morphology, its ability to ensure surface smoothness and fine-detail preservation is often limited. Therefore, the DL-generated geometry is used as an informed initialization for an iterative optimization module. By incorporating anatomical priors and explicit geometric constraints into the objective function, this module further optimizes mesh vertices to correct local distortions, improve boundary conformity, and maintain topological consistency, ultimately producing 3D geometries with both high anatomical fidelity and geometric quality.

4) Spatial registration: Finally, to ensure anatomical alignment with the source MRI, an inverse normalization procedure is performed. Using the archived scaling factors and rigid transformation parameters, the optimized geometry is registered back to the native spatial coordinate system of the original MRI, yielding the final anatomically aligned 3D geometry.

To delineate synergy between the DL and iterative optimization modules, we designed distinct implementation strategies for the training and inference phases.

During training, the framework adopts a "learning-from-optimization" strategy. For each training sample, the iterative optimization module performs an exhaustive search in the deformation space to identify an optimal displacement field. The resulting optimal displacement fields, obtained through computationally intensive optimization, are treated as pseudo-labels and used as supervisory signals to guide training of the DL model. The architecture of the proposed DL model is illustrated in Figure 2. The network employs a geometry-aware, multi-level architecture designed to learn a precise displacement field mapping. By integrating hierarchical feature fusion, the model captures both global shape contexts and fine-grained geometric signatures from unstructured point clouds to accurately reconstruct organ-specific morphologies.

During inference, the trained DL model provides an informed, hot-start initialization. Unlike conventional methods that use uninformed cold-start optimization from a generic spherical template, our model rapidly generates an initial displacement field closely matching the target anatomy from input point clouds. A lightweight iterative optimization module then refines the results, correcting residual errors to improve boundary fitting and surface smoothness.

### 2.3. Deep learning model

#### 2.3.1. Model architecture

The deep learning model consists of four main functional stages:

1) Dual-Branch graph construction: The network first constructs a graph by capturing local spatial relationships within the input point clouds using the K-nearest neighbor algorithm

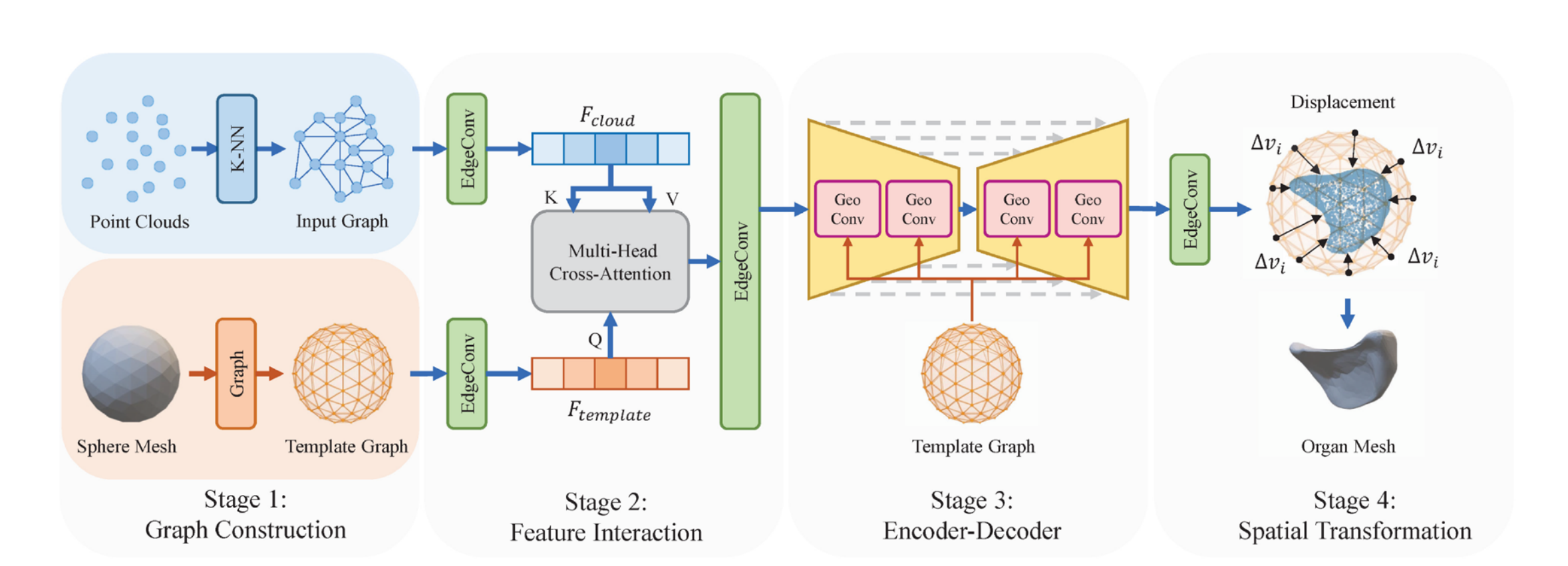

**Figure 2.** Architecture of the deep learning model for pelvic organ 3D geometric reconstruction. The model reconstructs high-fidelity organ meshes from segmentation-derived point clouds through four stages. Stage 1 constructs graphs from the input point clouds and the spherical template to encode local topology. Stage 2 uses multi-head cross-attention to fuse point-cloud features with template features. Stage 3 applies a geometry-aware encoder-decoder to learn hierarchical shape representations. Stage 4 predicts vertex-wise displacement vectors that deform the spherical template into the final patient-specific organ mesh.

(Cover and Hart 1967). In parallel, a predefined graph structure is established by leveraging the inherent vertex connectivity of the spherical template, enabling consistent topological correspondence during deformation learning.

2) Feature extraction and interaction: EdgeConv (Wang *et al* 2019) layers are employed to extract high-dimensional spatial embeddings from both the input point cloud graph and the template mesh graph. Subsequently, a multi-head cross-attention module (Vaswani *et al* 2017) is introduced to deeply fuse the dual-path features, in which the template features act as the Query (Q) and the point cloud features serve as the Key (K) and Value (V). This design is intended to explicitly model spatial correspondences between the canonical template and the patient-specific point cloud, thereby aiming to capture the morphological variation across anatomical structures. The fused features are then further processed through additional EdgeConv layers to effectively aggregate complex, nonlinear spatial contexts.

3) Geometry-aware encoder-decoder: To aggregate multi-scale geometric information, the network adopts a specialized encoder-decoder architecture integrated with geometry-aware convolutional layers. Inspired by the U-Net (Ronneberger *et al* 2015) hierarchical paradigm, this symmetric structure is designed to enable progressive feature abstraction and reconstruction. The intent of this configuration is to enhance the model's sensitivity to localized anatomical nuances, while preserving global topological stability and reducing non-physical mesh distortions during deformation.

4) Spatial transformation and displacement field output: Finally, the aggregated features are mapped through the terminal EdgeConv layer to generate a vertex-wise displacement field. By applying this displacement field to the original spherical template, the template is deformed into the target organ-specific 3D geometry, completing the mapping from a canonical sphere to the patient-specific anatomical structure.

### 2.3.2. Geometry-aware convolutional layer

To capture multi-scale geometric features and address the limitations of conventional 1D convolutional neural networks (1D Conv) (Kiranyaz *et al* 2021) when modeling intricate pelvic floor deformations, we developed a geometry-aware convolutional layer. This layer enhances model robustness to anatomical morphological variability by introducing explicit spatial priors. Specifically, the architecture employs a dynamic geometric weight generation mechanism that computes vertex-wise Euclidean distances, which are subsequently mapped to adaptive convolutional weights $w_{ij}^{geo}$ via a Gaussian radial basis function (RBF) (Hastie *et al* 2017):

$$w_{ij}^{geo} = \exp\left(-\frac{\| v_i - v_j \|_2^2}{2\sigma^2}\right) \tag{1}$$

where $\sigma$ denotes the hyperparameter regulating the bandwidth of the Gaussian Kernel. Building upon this formulation, we define a geometry-aware convolution operator. Given the input features of a neighboring node $\mathbf{x}_j \in \mathbb{R}^{C_{in}}$ and learnable kernel parameters $\mathbf{W} \in \mathbb{R}^{C_{out} \times C_{in}}$, the output feature $y_i$ is formulated as:

$$\mathbf{y}_i = \exp\left(-\frac{\| v_i - v_j \|_2^2}{2\sigma^2}\right) \sum_{j \in \mathcal{N}(i)} \mathbf{W}_j \cdot \left(w_{ij}^{geo} \cdot \mathbf{x}_j\right) + \mathbf{b} \tag{2}$$

In this configuration, $w_{ij}^{geo}$ acts as a dynamic spatial attention gate. By assigning higher importance to features from spatially proximal vertices, the operator adaptively calibrates the receptive field to local physical geometry, even under severe mesh compression or elongation. This mechanism helps suppress non-physical geometric distortions during deformation. Building upon this operator, we further extend the design into a multi-scale, U-Net-like hierarchical encoder-decoder architecture, enabling progressive mapping from global anatomical contours to local fine-scale geometric details through the fusion of deep contextual features and shallow geometric cues.

**2.4. Training strategy and loss functions**

Following the establishment of the network architecture, the primary challenge lies in enabling the model to effectively internalize complex geometric deformation patterns. To address this challenge, we designed a two-stage amortized optimization (Amos 2023) training strategy. The core rationale is to leverage data-driven learning to transform the originally time-consuming iterative geometric optimization process into a parameterized neural network mapping, thereby enabling rapid prediction of high-fidelity displacement fields during inference. The training procedure is divided into two distinct phases:

Stage 1: Supervised full-amortized learning. In this stage, the model performs a regression task analogous to expert demonstration. The optimal displacement fields obtained via iterative optimization serve as ground-truth supervision. The objective of this stage is to endow the network with fundamental shape reconstruction capability and to rapidly establish a global perception of the organ anatomy. The loss function is as follows:

$$\mathcal{L}_{first_stage} = \frac{1}{N}\sum_{i=1}^{N} |\hat{y}_i - y_i| + \frac{1}{N}\sum_{i=1}^{N} (\hat{y}_i - y_i)^2 \tag{3}$$

where $\hat{y}$ denotes the distance between the predicted displacement, and $y$ denotes the ground-truth displacement.

Stage 2: Objective-driven semi-amortized learning. After establishing baseline shape reconstruction capability, the model enters a geometry-aware fine-tuning stage driven by explicit geometric quality objectives. In this stage, training no longer relies on predefined displacement labels; instead, geometric quality loss functions (Wang *et al* 2021)—including edge length consistency, Laplacian smoothness, and normal consistency—are directly used as gradient supervision. The purpose of this stage is to leverage physical and geometric priors to correct residual prediction errors, thereby enhancing both boundary fidelity and surface quality of the reconstructed 3D geometry, particularly in regions with complex anatomical curvature. The loss function for this stage is defined as:

Edge Loss: Given an edge set $\varepsilon$ and vertex pairs $(v_i, v_j) \in \varepsilon$

$$\bar{l} = \frac{1}{|\mathcal{E}|} \sum_{(i,j)\in\mathcal{E}} \| v_i - v_j \|_2, \quad \mathcal{L}_{\text{edge}} = \sum_{(i,j)\in\mathcal{E}} \left(\| v_i - v_j \|_2 - \bar{l}\right)^2 \tag{4}$$

Laplacian Smoothness Loss: For each vertex $v_i$ with neighbors $\mathcal{N}(i)$

$$\mathcal{L}_{\text{lap}} = \sum_i \left\| v_i - \frac{1}{|\mathcal{N}(i)|} \sum_{j\in\mathcal{N}(i)} v_j \right\|_2^2 \tag{5}$$

Normal Loss: For predicted and ground truth face normals $n_i^{pred}$ and $n_i^{gt}$

$$\mathcal{L}_{\text{normal}} = \sum_i \left(1 - \left\langle n_i^{\text{pred}}, n_i^{\text{gt}} \right\rangle\right) \tag{6}$$

The total loss for the second stage, $\mathcal{L}_{secend_stage}$, is defined as a weighted combination of three loss terms:

$$\mathcal{L}_{seceond_stage} = \alpha\mathcal{L}_{edge} + \beta\mathcal{L}_{Laplacian} + \gamma\mathcal{L}_{normal} \tag{7}$$

where $\mathcal{L}_{edge}$ denotes the edge loss, $\mathcal{L}_{Laplacian}$ denotes the Laplacian smoothness loss, and $\mathcal{L}_{normal}$ denotes the normal loss. The chosen hyperparameters are $\alpha = 1.0, \beta = 0.1$ and $\gamma = 0.01$.

The underlying rationale of the two-stage training strategy is to temporally decouple global morphological learning from localized geometric refinement, thereby aiming to reduce optimization instability and minimize mutual interference between coarse shape convergence and fine-scale feature fitting during early stages of training.

**2.5. Evaluation metrics**

To evaluate the proposed framework, geometric fidelity—defined as the fidelity of anatomical restoration—was measured via Chamfer Distance (CD) and Dice Similarity Coefficient (DSC), while geometric quality—defined as the preservation of surface details and smoothness—was assessed via Normal Consistency (NC), Edge Loss Consistency (ELC), and Laplacian Smoothing (LS). CD measures the spatial alignment between point clouds sampled from the reconstructed geometric surface and the reference surface, with lower CD values indicating closer geometric correspondence. To ensure an unbiased and consistent evaluation, 5,000 points are uniformly sampled from each 3D surface, thereby maintaining a standardized sampling density across all models. CD is defined as:

$$CD(P,Q) = \frac{1}{|P|}\sum_{p\in P} \min_{q\in Q} \|p - q\|_2^2 + \frac{1}{|Q|}\sum_{q\in Q} \min_{p\in P} \|q - p\|_2^2 \tag{8}$$

where $P$ and $Q$ denote the predicted and ground-truth point sets, respectively, while $|P|$ and $|Q|$ denote the number of points in each set. The variables $p$ and $q$ refer to individual 3D points within sets $P$ and $Q$.

DSC measures the volumetric overlap between the reconstructed geometry and the ground truth, with higher values indicating better spatial correspondence. For implementation, the reconstructed and reference 3D models are discretized into a $64^3$ voxel grid, enabling quantitative evaluation of their volumetric overlap. DSC is computed as follows:

$$\text{DSC} = \frac{2|\hat{Y} \cap Y|}{|\hat{Y}| + |Y|} = \frac{2\sum_i \hat{y}_i\, y_i}{\sum_i \hat{y}_i + \sum_i y_i}. \tag{9}$$

where $\hat{Y}$ and $Y$ denote the sets of predicted and ground-truth occupied voxels, respectively, where $|\cdot|$ represents the cardinality of each set. The variable $\hat{y_i}$ refers to the predicted occupancy of the $i$-th voxel, while $y_i$ denotes its the ground-truth label.

NC quantifies the local geometric smoothness and structural consistency of the reconstructed surface, with lower values indicating smoother surfaces and fewer local geometric artifacts under the current definition. NC is defined as:

$$NC = \frac{1}{|E|}\sum_{e \in E}\left(1 - \frac{n_0 \cdot n_1}{\| n_0 \| \| n_1 \|}\right) \tag{10}$$

where $|E|$ denotes the total number of shared edges in the mesh that are incident to at least two faces, and $e \in E$ represents an individual shared edge. The variables $n_0$ and $n_1$ represent the unit normal vectors of the two adjacent faces incident to edge $e$, respectively.

LS measures the local structural coherence between mesh vertices and their neighboring vertices, with lower values indicating smoother surfaces and more physically plausible local deformations. LS is defined as follows:

$$LS = \sum_{p} \| \delta_p' - \delta_p \|_2^2 \tag{11}$$

where,

$$\delta_p = p - \sum_{k \in \mathcal{N}(p)} \frac{1}{\| \mathcal{N}(p) \|} k \tag{12}$$

where, for any vertex $p$, the Laplacian coordinate $\delta_p$ is defined as the vector difference between the vertex and the centroid of its 1-ring neighborhood. $\mathcal{N}(p)$ denotes the set of adjacent vertices $k$ directly connected to $p$. The variables $\delta_p'$ and $\delta_p$ correspond to the Laplacian coordinates of the vertex after and before deformation.

ELC evaluates the uniformity and physical plausibility of the mesh topology, with lower values indicating a more balanced distribution of edge lengths and reduced occurrence of extreme edge stretching or compression.

$$ELC = \sum_{p} \sum_{k \in \mathcal{N}(p)} \| p - k \|_2^2 \tag{13}$$

The definitions of all variables are consistent with those in Laplacian smoothness.

To facilitate a comprehensive comparison across dimensions with disparate units and optimization directions, we introduced a Normalized Performance Score ($S_j$) for each individual metric $j$. This metric unifies all indices into a "higher is better" logic by mapping the best observed performance in the experimental cohort to a 100% baseline. Specifically, for a given method $i$, the normalized score for each metric is defined as:

$$S_{i,j} = \begin{cases} \frac{V_{min,j}}{V_{i,j}} \times 100\% & \text{for } j \in \{\text{CD}, \text{NC}, \text{ELC}, \text{LS}\} \\ \frac{V_{i,j}}{V_{max,j}} \times 100\% & \text{for } j \in \{\text{DSC}\} \end{cases} \tag{14}$$

where $V_{i,j}$ represents the observed value of metric $j$ for method $i$. $V_{min,j}$ and $V_{max,j}$ denote the best observed values within the study for each respective metric. To evaluate the overall trade-off between anatomical fidelity and geometric stability, the Weighted Overall Score is then computed as the weighted sum of these individual scores:

$$S_{overall} = \sum_{j \in \{\text{CD, DSC, ELC, NC, LS}\}} w_j S_{i,j} \tag{15}$$

The weights $w_j$ were assigned based on hierarchical priority: $w_{CD} = w_{DSC} = 0.35$ for anatomical fidelity, and $w_{NC} = w_{ELC} = w_{LS} = 0.1$ for geometric quality. This weighting scheme reflects the principle that anatomical fidelity is the fundamental prerequisite for application; any geometric refinement (e.g., surface smoothness) is secondary and meaningful only when structural fidelity is maintained.

**2.6. Comparison methods**

To evaluate the performance of the DL backbone, we conducted a head-to-head comparison between our model and representative mainstream DL approaches for 3D organ geometric reconstruction, including Voxel2Mesh (Wickramasinghe *et al* 2020), MeshDeformNet (Kong *et al* 2021), and DeepCarve (Pak *et al* 2024). For fair comparison, all methods were provided with manually annotated expert contours as segmentation masks in all experiments. Subsequently, to investigate the coupling mechanism between the DL and iterative optimization modules, we compared a single-path strategy of direct optimization from the original sphere template with a hybrid strategy that combines DL-based hot-start initialization followed by iterative optimization.

To assess the usability of our framework for downstream biomechanical modeling, we benchmarked the reconstructed geometries against conventional post-processing algorithms, including Laplacian, Taubin, and Humphrey smoothing. Given that finite element simulations require well-defined volumetric elements (Sbriglio *et al* 2026), the reconstructed surface geometries were further converted into computable volumetric mesh models to support downstream biomechanical analysis. Specifically, we employed Gmsh (Geuzaine and Remacle 2009) in conjunction with the 3D Delaunay algorithm (Watson 1981) to generate tetrahedral meshes from the reconstructed 3D geometries.

To evaluate the numerical stability and mesh-quality characteristics of the generated 3D geometries, we employed a set of well-established element-wise metrics commonly utilized within the finite element analysis (FEA) community: minSICN (minimal signed inverted condition number), minSIGE (minimal signed inverted gradient error), and Gamma (the ratio of the inscribed to circumscribed sphere radius). These standard benchmarks are specifically chosen for their effectiveness in detecting element inversions and excessive geometric distortions—factors that frequently lead to computational non-convergence in large-deformation biomechanical simulations (Geuzaine and Remacle 2009, Knupp 2001, Shewchuk 2002). Furthermore, as the robustness of FEA is often governed by the "bottleneck" of poor-quality elements, we assessed each sample based on both the lowest element-wise value and the average of the 10 worst-quality elements. This rigorous approach ensures that the reconstructed geometries are suitable for downstream high-fidelity biomechanical applications.

**2.7. Implementation details**

To evaluate the effectiveness of the proposed method, we adopted a randomized patient-level data splitting strategy. The complete dataset (100 participants) was partitioned into a training set (75 participants, 225 organs), a validation set (5 participants, 15 organs), and an independent test set (20 participants, 60 organs). The test set was strictly isolated from the training and validation processes to preclude any potential data leakage.

To facilitate reproducibility, we provide the following details: Throughout the training process, we conducted on four NVIDIA A5000 graphics cards equipped with 24GB of computing memory, using the Adam optimizer to accelerate model training convergence; all models were trained for 500 epochs, batch size 8, learning rate 0.0001.

For statistical analysis, we conducted 2-tailed paired t tests for comparison. All indicators are reported as mean values and 95% confidence intervals (CI), and $P < 0.05$ was considered statistically significant. Statistical analyses were performed using IBM SPSS (version 23).

## 3. Results

### 3.1. Deep learning results

Table 1 shows the quantitative comparison results of the proposed DL model and mainstream DL models for 3D organ geometric reconstruction in pelvic floor organ reconstruction tasks. The results demonstrate that our model consistently showed better

**Table 1.** CD and DSC comparison with other deep learning models for 3D organ geometric reconstruction.

| Organ | Methods | CD( ↓ ) | P-value | DSC( ↑ ) | P-value |
|---|---|---|---|---|---|
| Bladder | Voxel2Mesh | 0.0312** (0.0220-0.0404) | <0.01 | 0.586** (0.5364-0.6356) | <0.01 |
| | MeshDeformNet | 0.0280** (0.0218-0.0342) | <0.01 | 0.616** (0.5725–0.6595) | <0.01 |
| | DeepCarve | 0.0969** (0.0847–0.1091) | <0.01 | 0.331** (0.2753–0.3867) | <0.01 |
| | Ours | **0.0055 (0.0048–0.0062)** | -- | **0.788 (0.7506–0.8254)** | -- |
| Rectum | Voxel2Mesh | 0.0258** (0.0148–0.0368) | <0.01 | 0.599** (0.5602–0.6378) | <0.01 |
| | MeshDeformNet | 0.0230** (0.0097–0.0363) | <0.01 | 0.501** (0.4547–0.5473) | <0.01 |
| | DeepCarve | 0.1725** (0.1505–0.1945) | <0.01 | 0.178** (0.1167–0.2393) | <0.01 |
| | Ours | **0.0037 (0.0027–0.0047)** | -- | **0.839 (0.8189–0.8591)** | -- |
| Uterus | Voxel2Mesh | 0.0298** (0.0195–0.0401) | <0.01 | 0.726** (0.6806–0.7714) | <0.01 |
| | MeshDeformNet | 0.0255** (0.0204–0.0306) | <0.01 | 0.740** (0.7124–0.7676) | <0.01 |
| | DeepCarve | 0.0968** (0.0834–0.1102) | <0.01 | 0.446** (0.4006–0.4914) | <0.01 |
| | Ours | **0.0042 (0.0036–0.0048)** | -- | **0.900 (0.8930–0.9070)** | -- |

*Note:* The best performance (mean value) is highlighted in bold. The numbers in the parentheses represent the 95% confidence intervals (95% CI). Statistical significance was assessed using a two-tailed paired t-test.

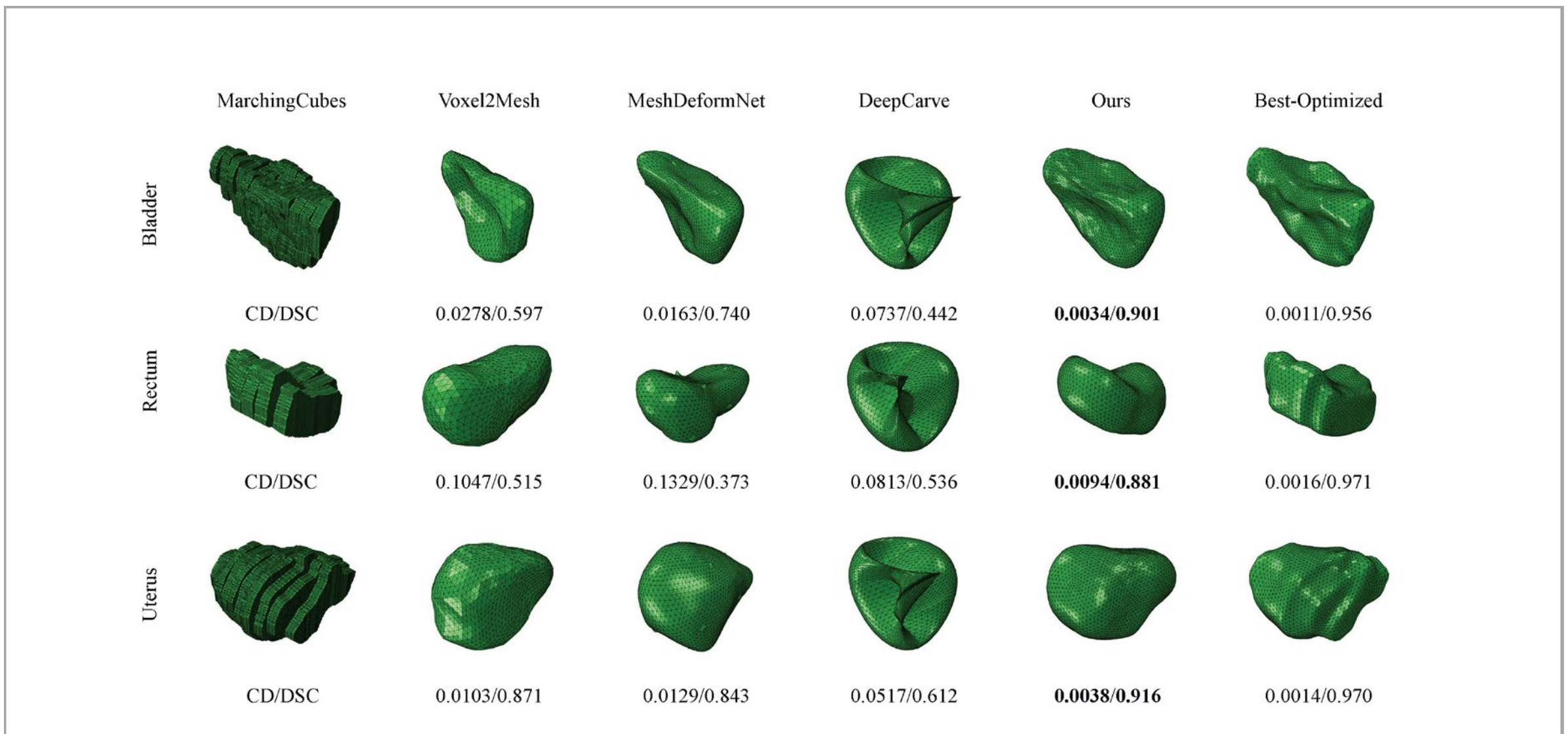

**Figure 3.** Qualitative comparison of 3D pelvic organ geometries reconstructed by different deep learning models. From left to right, the columns show the initial geometries generated by Marching Cubes, the results from Voxel2Mesh, MeshDeformNet, and DeepCarve, the results from our model, and the optimized geometric reference. Representative examples are shown for the bladder, rectum, and uterus.

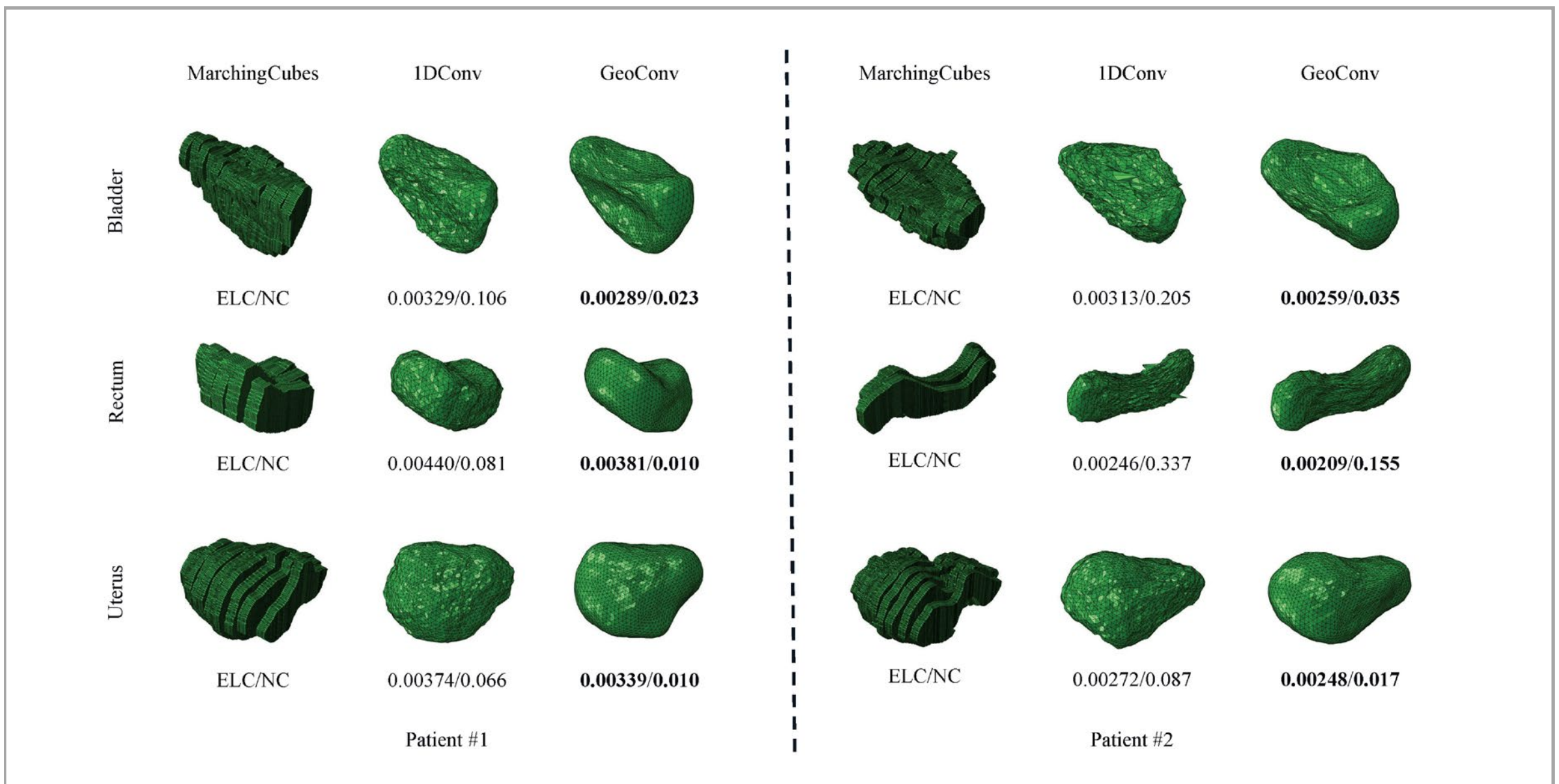

**Figure 4.** Qualitative comparison for 3D reconstruction outcomes between two convolutional architectures using representative cases from the test cohort. From left to right, the columns show (1) Marching Cubes; (2) 1D convolutional layer (1DConv); and (3) the geometry-aware convolutional layer (GeoConv).

performance across all geometric fidelity metrics. Notably, regarding the CD, our model achieved an order-of-magnitude reduction in reconstruction error compared with baseline models. Statistical analysis using a two-tailed paired t-test confirmed that these performance improvements in the geometric fidelity metrics were statistically significant ($P<0.05$) across all comparative benchmarks.

Complementing these quantitative findings, Figure 3 illustrates the qualitative visual performance for the three pelvic organs. Visual inspection suggests the marked superiority of our framework in morphological restoration. Specifically, the reconstructed geometries exhibit high-fidelity alignment with the initial geometries (derived via the Marching Cubes algorithm) in terms of macroscopic topology. Furthermore, our model better captures fine anatomical details, including sharp curvature changes and complex organ boundary transitions, yielding results close to the best-optimized geometries. By contrast, other comparison methods often suffer from severe geometric distortion and loss of key anatomical features under extreme deformations, failing to faithfully represent complex pelvic structures.

### 3.2. Ablation study

#### 3.2.1. Geometry-aware convolutional layer

Figure 4 illustrates a qualitative comparison of the 3D reconstruction outcomes between the two convolutional architectures using representative cases from the test cohort. It should be noted that this comparison is based on the outputs from the first stage of training. Qualitative analysis suggests that, compared with standard convolutions, the geometry-aware convolution reduces surface roughness and suppresses non-physical geometric noise arising from insufficient spatial feature characterization.

Table 2 provides a detailed quantitative performance evaluation. This evaluation is also based on the first-stage training outputs. The results indicate that both architectures yield comparable performance regarding geometric fidelity. However, a clear divergence emerges in the metrics evaluating geometric quality. The geometry-aware convolution consistently outperformed the standard convolution across all three geometric quality dimensions. Most notably, the geometry-aware convolution achieved an order-of-magnitude reduction in error metrics associated with normal consistency and Laplacian smoothness.

#### 3.2.2. Semi-amortization

Figure 5 provides a comprehensive visual comparison of different training strategies for the pelvic organ reconstruction task. As demonstrated by the Normalized Performance Scores, our proposed two-stage training strategy achieves a superior overall balance between anatomical fidelity and geometric quality. Quantitative analysis across multiple dimensions suggests that full-amortization does not maintain surface integrity, leading to reduced geometric quality. Conversely, semi-amortization enhances geometric smoothness but at the considerable expense of reconstruction fidelity. In contrast, the two-stage training strategy introduced in this study is designed to achieve high anatomical fidelity while simultaneously maintaining geometric quality at a level comparable to the semi-amortization approach. This pattern is further reflected by the highest Overall Score, suggesting that prioritizing structural fidelity while progressively refining geometric stability provides a strong solution for mesh reconstruction.

#### 3.2.3. Two-stage hybrid framework

Figure 6 illustrates the convergence curves of the DSC with respect to the number of iterations for the three target pelvic organs under the two different inference pathways. Regarding convergence trajectories, the two-stage hybrid framework and the standalone optimization pathway exhibit a pronounced disparity in computational efficiency. Specifically, the hybrid framework attains a steady-state plateau within approximately 100 iterations (approximately 15 seconds), whereas the standalone optimization pathway demonstrates a persistent upward trend at the same number of iterations, indicating slower convergence. Even after reaching 1,000 iterations (approximately 150 seconds), the mean DSC achieved by the hybrid framework across all organs consistently surpassed that of the pure optimization control group.

**Table 2.** Model performance comparison using different convolutional layers.

| Organ | Methods | CD (↓) | P-value | DSC(↑) | P-value | ELC(↓) | P-value | NC(↓) | P-value | LS(↓) | P-value |
|---|---|---|---|---|---|---|---|---|---|---|---|
| Bladder | 1DConv | **0.00681 (0.00572–0.00790)** | 0.772 | 0.769 (0.730–0.808) | 0.768 | 0.00354** (0.00330–0.00378) | <0.01 | 0.215** (0.176–0.254) | <0.01 | 0.0179** (0.0166–0.0192) | <0.01 |
| | GeoConv | 0.00681 (0.00567–0.00795) | -- | **0.773 (0.733–0.813)** | -- | **0.00280 (0.00263–0.00297)** | -- | **0.035 (0.026–0.0439)** | -- | **0.0069 (0.0066–0.0072)** | -- |
| Rectum | 1DConv | **0.00439 (0.00356–0.00522)** | 0.142 | **0.818 (0.791–0.845)** | 0.269 | 0.00276** (0.00253–0.00299) | <0.01 | 0.243** (0.191–0.295) | <0.01 | 0.0145** (0.0134–0.0256) | <0.01 |
| | GeoConv | 0.00488 (0.00377–0.00599) | -- | 0.812 (0.786–0.838) | -- | **0.00235 (0.00213–0.00257)** | -- | **0.076 (0.056–0.096)** | -- | **0.0067 (0.0064–0.0070)** | -- |
| Uterus | 1DConv | 0.00596 (0.00500–0.00692) | 0.156 | 0.876 (0.866–0.885) | 0.026 | 0.00330** (0.00305–0.00355) | <0.01 | 0.136** (0.111–0.161) | <0.01 | 0.0149** (0.0141–0.0157) | <0.01 |
| | GeoConv | **0.00561 (0.00468–0.00654)** | -- | **0.884 (0.875–0.892)** | -- | **0.00284 (0.00260–0.00308)** | -- | **0.018 (0.013–0.023)** | -- | **0.0059 (0.0057–0.0061)** | -- |

*Note:* The best performance (mean value) is highlighted in bold. The numbers in the parentheses represent the 95% confidence intervals (95% CI). Statistical significance was assessed using a two-tailed paired t-test. 1DConv: 1D convolutional layer; GeoConv: geometry-aware convolutional layer.

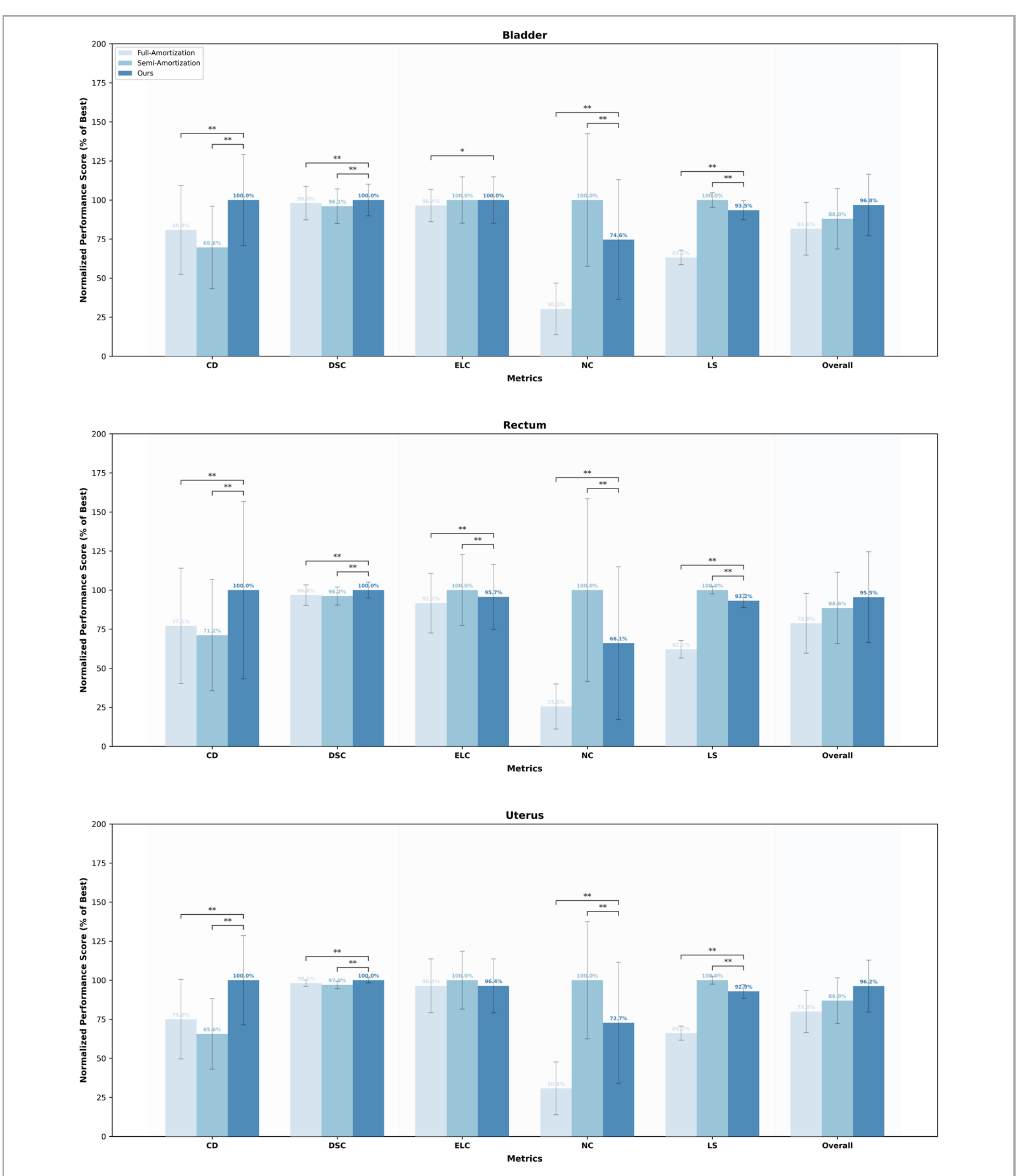

**Figure 5.** Comparative performance of training strategies across three pelvic organs. The bars illustrate the performance of Full-Amortization, Semi-Amortization, and the proposed model (Ours) for Bladder, Rectum, and Uterus. Performance is evaluated across six dimensions: (1) fidelity: Chamfer Distance (CD) and Dice Similarity Coefficient (DSC); (2) geometric quality: Normal Consistency (NC), Edge Loss Consistency (ELC), and Laplacian Smoothing (LS); and (3) overall: a weighted composite score. All results are aggregated from 20 subjects. Metrics are normalized to a percentage of the best-observed mean value (100% denotes the best observed result in each category). Data are presented as mean values (labels) with error bars denoting standard deviation (SD). Significance levels are denoted as * ($P < 0.05$) and ** ($P < 0.01$) to indicate statistically significant differences compared with our proposed model. The Overall score is provided as a composite performance index for descriptive comparison and was not subjected to separate statistical inference.

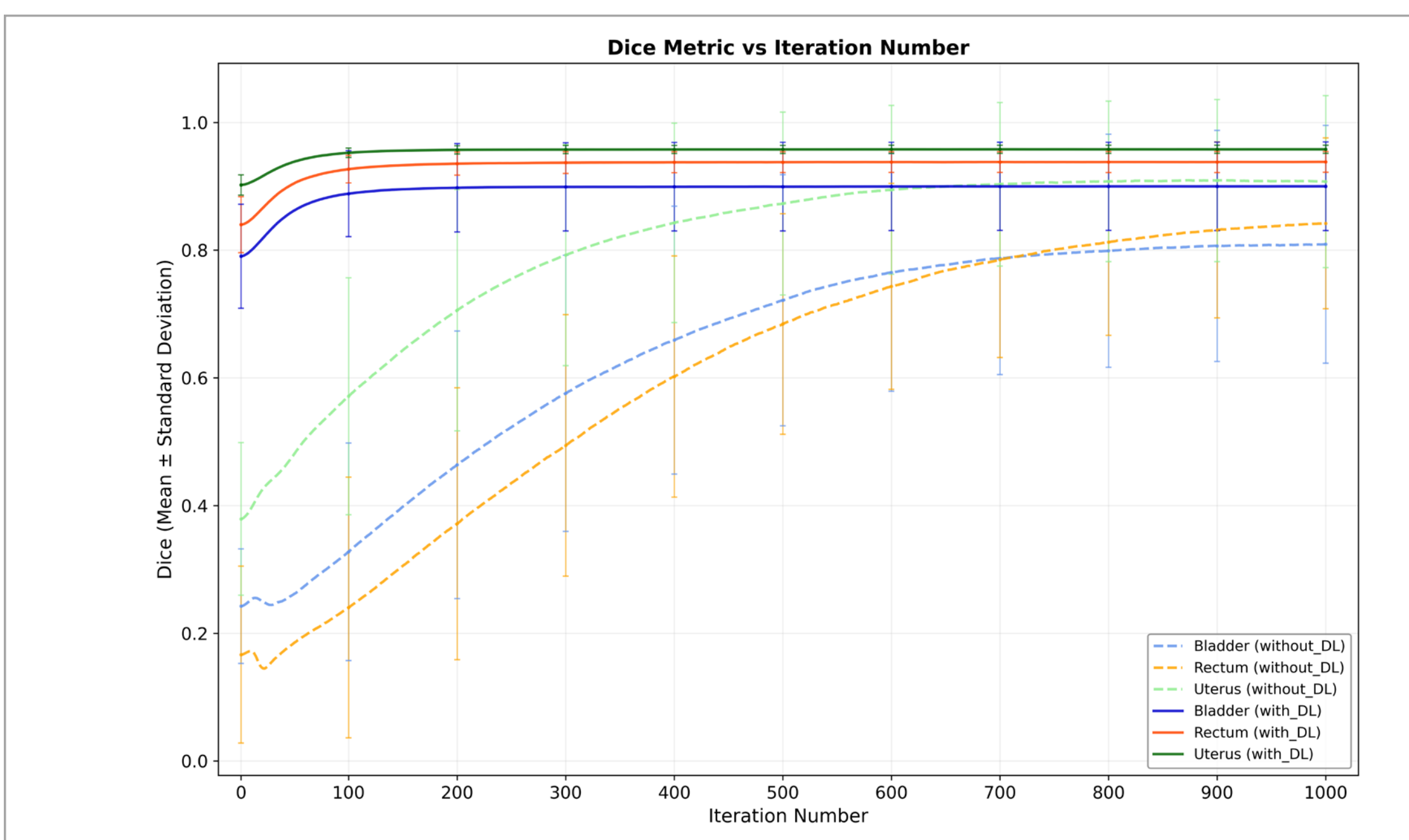

**Figure 6.** Comparison of Dice metric convergence over iteration numbers for different pelvic organs with and without deep learning assistance. The curves show the geometry reconstruction performance for the bladder (blue), rectum (orange), and uterus (green). Dashed lines represent direct optimization without deep learning initialization, whereas solid lines represent the hybrid pathway with deep learning hot-start initialization. Error bars represent the mean and standard deviation.

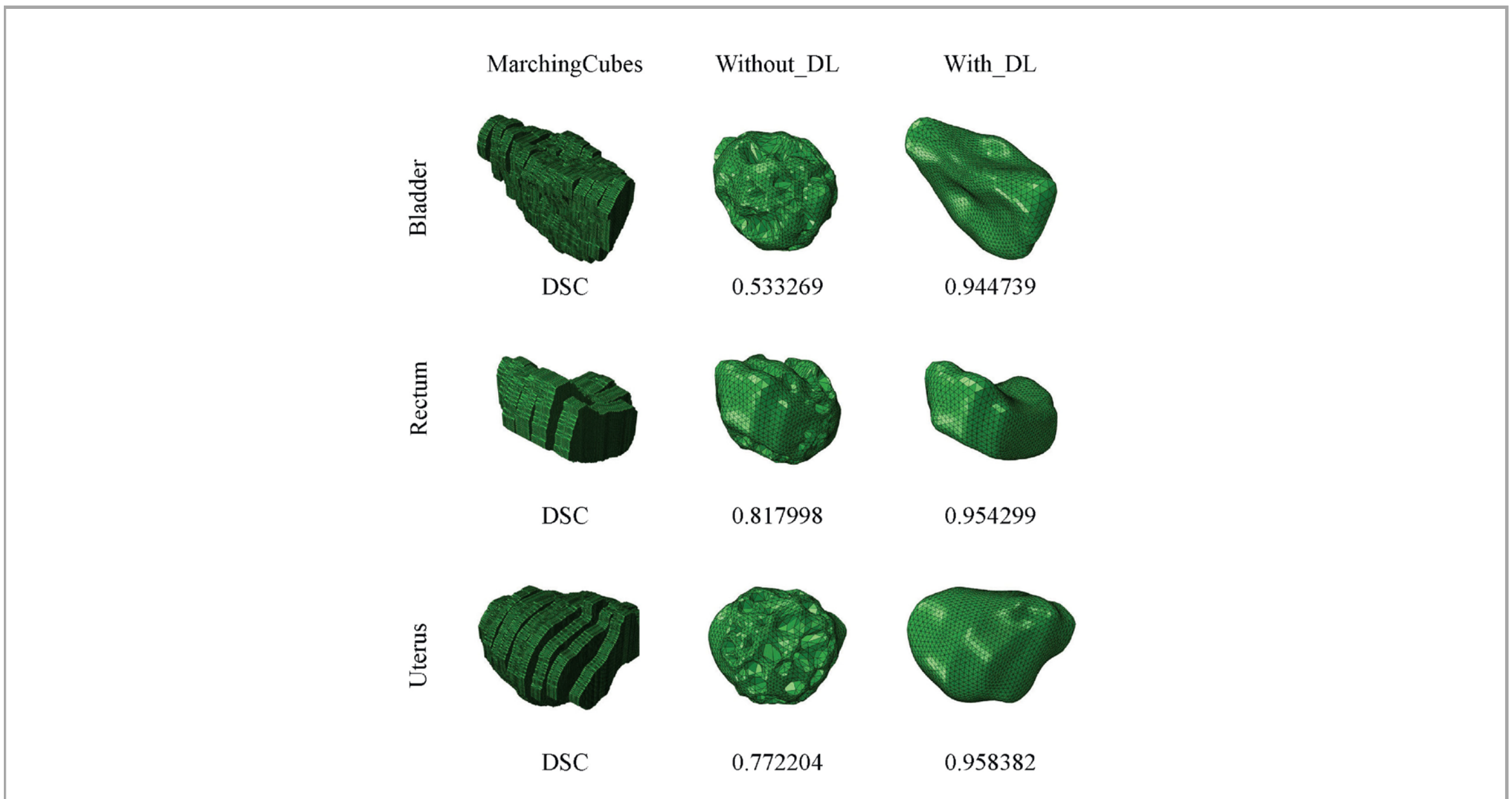

**Figure 7.** Visual comparison of geometric outcomes at 100 iterations with and without deep learning assistance. From left to right, the columns show the initial mesh generated by Marching Cubes, the result after 100 iterations of direct optimization without deep learning assistance, and the result after 100 iterations with deep learning assistance. Representative examples are shown for the bladder, rectum, and uterus, and the values below each reconstructed geometry indicate the Dice similarity coefficient.

Furthermore, the qualitative comparisons presented in Figure 7 highlight the distinct geometric configurations achieved by the two methods at the 100-iteration checkpoint. At this stage, the two-stage hybrid framework has already generated anatomically plausible geometries with high fidelity, whereas the geometries produced by the standalone optimization pathway remain characterized by primitive contours and fail to achieve effective morphological convergence toward the target anatomical structures within the same timeframe.

### 3.3. Compared with traditional operators

**Table 3.** Performance comparison with different 3D post-processing algorithms.

| Statistics | Methods | minSICN(↑) | P-value | minSIGE(↑) | P-value | Gamma(↑) | P-value |
|---|---|---|---|---|---|---|---|
| Lowest Element-wise Value | Laplacian | 0.0095** (0.0062–0.0128) | <0.01 | 0.0112* (0.0071–0.0153) | <0.05 | 0.00234 (0.00013–0.00455) | 0.052 |
| | Humphrey | 0.0028** (0.0021–0.0035) | <0.01 | 0.0028** (0.0021–0.0036) | <0.01 | 0.00005* (0.00003–0.00007) | <0.05 |
| | Taubin | 0.0097** (0.0064–0.013) | <0.01 | 0.0113** (0.0072–0.0154) | <0.05 | 0.00237 (0.00016–0.00458) | 0.052 |
| | Ours | **0.0409 (0.0303–0.0515)** | -- | **0.0416 (0.0309–0.0523)** | -- | **0.02275 (0.01200–0.03350)** | -- |
| Mean of 10 Worst Elements | Laplacian | 0.0247** (0.0178–0.0316) | <0.01 | 0.0274** (0.0199–0.0349) | <0.01 | 0.01105* (0.00513–0.01697) | <0.05 |
| | Humphrey | 0.0081** (0.0069–0.0093) | <0.01 | 0.0082** (0.0070–0.0094) | <0.01 | 0.00023** (0.00016–0.00030) | <0.01 |
| | Taubin | 0.0261** (0.0193–0.033) | <0.01 | 0.0289** (0.0214–0.0364) | <0.01 | 0.01185* (0.00596–0.01774) | <0.05 |
| | Ours | **0.0807 (0.0672–0.0942)** | -- | **0.0844 (0.0706–0.0982)** | -- | **0.04921 (0.03417–0.06424)** | -- |

*Note:* The best performance (mean value) is highlighted in bold. The numbers in the parentheses represent the 95% confidence intervals (95% CI). Statistical significance was assessed using a two-tailed paired t-test.

The quantitative metrics in Table 3 indicate that our framework improves tetrahedral mesh quality under the evaluated settings. Compared with traditional operators like Laplacian and Taubin smoothing, our approach achieves a multi-fold increase in the mesh quality lower bound. Unlike these traditional operators that risk mesh shrinkage, our geometric module proactively optimizes element distribution based on the learned manifold structure. Consequently, our framework effectively elevates both the Lowest Element-wise Value and the Mean of 10 Worst Elements, indicating improved local element quality during large-deformation analysis.

## 4. Discussion

This study presents a hybrid deformable shape modeling framework that combines DL prediction with iterative optimization for high-fidelity 3D geometric reconstruction of pelvic organs from pelvic MRI. The approach offers three key innovations. First, the framework employs a geometry-aware multi-level deep learning architecture that fuses dual-branch graph features via geometric convolution and cross-attention, preserving topological consistency while capturing global and fine-grained geometry from point clouds. Second, a two-stage amortized optimization training strategy is introduced, which first learns global shapes through supervised full amortization and then refines local surfaces via semi-amortization, temporally decoupling reconstruction fidelity and physical constraints to enhance training stability. Third, a synergistic mechanism is established between deep learning and iterative optimization: during training, iterative optimization supervises deep learning through pseudo-labels; during inference, deep learning provides a hot-start initialization to guide rapid iterative deformation, thereby achieving both high fidelity and computational efficiency.

As demonstrated in Table 1 and Figure 3, our model achieves significantly higher geometric fidelity compared to existing approaches. This leap stems from our end-to-end geometric DL architecture, which moves away from conventional image-point cloud multimodal fusion in favor of a streamlined, manifold-oriented pipeline. In the 3D imaging domain, raw voxel data is high-dimensional yet carries a very low density of information relevant to organ surface geometry. This low signal-to-noise ratio often causes models trained on small datasets to capture irrelevant background noise, leading to severe overfitting.

According to the quantitative evaluation in Table 2, the 3D geometric models generated using geometry-aware convolution are significantly superior to those generated by traditional convolution in terms of mesh quality indicators. From a modeling perspective, while standard 1D convolutions can effectively approximate global spatial coordinates—ensuring macroscopic anatomical accuracy—their lack of explicit spatial priors renders them prone to generating non-physical discrete noise and distortions on the reconstructed surfaces. In contrast, our proposed geometry-aware convolution operator is engineered to adaptively perceive the Euclidean distance between vertices, dynamically recalibrating the receptive field. This mechanism significantly bolsters surface smoothness and manifold integrity while maintaining high reconstruction precision.

As demonstrated in Figure 5, our proposed two-stage amortized optimization training strategy, viewed through the lens of curriculum learning, achieves superior overall performance by systematically decoupling conflicting gradient signals. In conventional training frameworks, loss terms for anatomical accuracy and those for surface regularization often produce mutually opposing gradients, creating an optimization dilemma between reconstruction fidelity and mesh integrity. Our curriculum-based strategy addresses this by temporally separating these objectives: the model first learns a globally coherent anatomical topology driven by reconstruction objectives. Then, in the second stage, the emphasis on reconstruction performance is set aside, and regularization constraints are introduced to refine the mesh manifold. This staged learning progression effectively disentangles gradient conflicts, allowing the model to bypass energy barriers caused by poor initial geometries and achieve both high-fidelity anatomical restoration and well-regularized mesh surfaces.

The quantitative metrics in Table 3 underscore our framework's advantages in refining tetrahedral meshes for downstream biomechanical workflows. Compared to traditional operators like Laplacian and Taubin smoothing, our approach achieves a multi-fold increase in the mesh quality lower bound. Unlike these traditional operators that risk mesh shrinkage, our geometric module proactively optimizes element distribution based on the learned manifold structure. Consequently, our framework effectively elevates both the Minimum Value and the Mean of 10 Worst Elements, indicating improved local element quality under the evaluated settings.

This study provides a new perspective for patient-specific pelvic floor modeling in clinically relevant research settings. While current research mostly focuses on voxel-wise segmentation (Wang *et al* 2025, Ma *et al* 2024), these methods often neglect topological continuity and geometric integrity, leaving a labor-intensive "technical gap" in post-processing that hinders sophisticated downstream analysis. Our proposed framework bridges this gap by facilitating automated 3D geometric reconstruction directly from segmentation outputs, reducing manual mesh refinement burden and improving the fidelity and efficiency of downstream patient-specific workflows (Peng *et al* 2015, Silva *et al* 2021, Mayeur *et al* 2016). For example, the high-quality, topologically consistent 3D geometries produced by our framework can serve as suitable inputs for downstream biomechanical finite element analysis—an approach widely used to investigate patient-specific structural damage mechanisms in pelvic floor disorders (Luo *et al* 2014, 2023) and to support the computational evaluation of pelvic floor-related medical devices (Craven *et al* 2025). By providing a standardized and automated reconstruction pipeline, our framework reduces the manual effort and inter-operator variability that currently limit the scalability of such analyses in both clinical and research settings. From a clinical perspective, our framework establishes a standardized anatomical baseline by employing a unified topological template for all reconstructions. This approach effectively eliminates the structural inconsistencies and variable vertex counts prevalent in traditional modeling, ensuring that all patient-specific models share a precise, point-to-point anatomical correspondence through homeomorphic mapping. Such "spatial normalization" enables clinicians to perform robust, large-scale statistical shape analysis and reliable inter-individual morphological comparisons that were previously hindered by technical gaps. By providing a uniform digital benchmark for pelvic organs, our method not only enhances geometric fidelity but also may support population-based research and the longitudinal monitoring of disease progression. Moving forward, we plan to validate these outcomes across a wider spectrum of patient cohorts and clinical environments to further assess the stability and generalizability of the model in diverse real-world scenarios.

The primary strength of this study is the proposed hybrid deformable shape modeling framework combining DL and iterative optimization modules, which effectively addresses the time consuming, labor intensive, and difficult to standardize challenges of reconstructing high quality 3D pelvic organ geometries based on pelvic MRI segmentation. However, there are prerequisites and challenges to consider. First, the efficacy of current modeling approaches remains highly dependent on the accuracy of input segmentation labels or trained segmentation models; achieving end-to-end direct mapping from raw MRI images to 3D geometric representations remains a key direction for future improvement. Second, the generalization and robustness of the model when handling multi-center and cross-device imaging data still require further validation with large-scale, heterogeneous clinical cohorts. Finally, the current deformation framework is primarily limited to organ geometries with sphere-like manifolds; moreover, while the current geometric indices indicate improved mesh-quality proxy performance, future studies will incorporate specific boundary conditions to evaluate the mechanical response in dynamic physiological scenarios.

## 5. Conclusions

In this study, we developed a hybrid deformable shape modeling framework that integrates DL-based prediction with iterative optimization to efficiently reconstruct high-fidelity 3D geometries of pelvic organs. By combining a geometry-aware, multi-level DL architecture, the framework predicts displacement fields from a spherical template to complex organ morphologies, followed by targeted iterative refinement of local geometric details. Experimental results demonstrate that the proposed approach achieves both high computational efficiency and superior geometric fidelity, establishing a robust technical foundation for future patient-specific pelvic floor modeling and analysis.

## Acknowledgements

This study was supported by the National Key R&D Program of China (grant no. 2023YFC2411201); NSFC General Program (grant no. 31870942); Peking University Clinical Medicine Plus X - Young Scholars Project (grant nos. PKU2020LCXQ017 and PKU2021LCXQ028); PKU-Baidu Fund (grant no. 2020BD039).

## Data availability statement

The data cannot be made publicly available upon publication because they contain sensitive personal information. The data that support the findings of this study are available upon reasonable request from the authors.

## Conflict of interest

The authors declare that they have no known competing financial interests or personal relationships that could have appeared to influence the work reported in this paper.

## Author contributions

Hui Wang
Conceptualization (equal), Methodology (equal), Data analysis (equal), Writing – original draft (equal), Writing – review and editing (equal).

Xiaowei Li
Data curation (equal), Data analysis (equal).

Chenxin Zhang
Data analysis (equal).

Yifan Feng
Data curation (equal).

Jianwei Zuo
Data analysis (equal).

Yumeng Tang
Data analysis (equal).

Xiuli Sun
Data curation (equal).

Jianliu Wang
Methodology (equal), Conceptualization (equal).

Bing Xie
Data curation (lead), Data analysis (equal).

Jiajia Luo
Conceptualization(lead), Methodology(lead), Data analysis(lead), Writing – original draft (equal), Writing – review and editing (equal).